\documentclass{article}

\usepackage{arxiv}

\usepackage[utf8]{inputenc} 
\usepackage[T1]{fontenc}    
\usepackage{hyperref}       
\usepackage{url}            
\usepackage{booktabs}       
\usepackage{amsfonts}       
\usepackage{nicefrac}       
\usepackage{microtype}      
\usepackage{lipsum}
\usepackage{graphicx}
\graphicspath{ {./images/} }
\usepackage{amsmath,amssymb,amsfonts}
\usepackage{multirow}

\title{Discrete Wavelet Transform as a Facilitator for Expressive Latent Space Representation in Variational Autoencoders in Satellite Imagery}

\author{%
\begin{tabular}{ccc}
\normalfont Arpan Mahara & \normalfont Md Rezaul Karim Khan & \normalfont Naphtali Rishe \\
\texttt{amaha038@fiu.edu} & \texttt{mkhan157@fiu.edu} & \texttt{rishen@cs.fiu.edu}
\end{tabular}
\\[0.8em]
\begin{tabular}{cc}
\normalfont Wenjia Wang & \normalfont Seyed Masoud Sadjadi \\
\texttt{wwang048@fiu.edu} & \texttt{sadjadi@cs.fiu.edu}
\end{tabular}
\\[1.0em]
\normalfont Knight Foundation School of Computing and Information Sciences \\
\normalfont Florida International University \\
\normalfont Miami, FL 33199
}

\begin{document}
\maketitle
\begin{abstract}
Latent Diffusion Models (LDM), a subclass of diffusion models, mitigate the computational complexity of pixel-space diffusion by operating within a compressed latent space constructed by Variational Autoencoders (VAEs), demonstrating significant advantages in Remote Sensing (RS) applications. Though numerous studies enhancing LDMs have been conducted, investigations explicitly targeting improvements within the intrinsic latent space remain scarce. This paper proposes an innovative perspective, utilizing the Discrete Wavelet Transform (DWT) to enhance the VAE’s latent space representation, designed for satellite imagery. The proposed method, ExpDWT-VAE, introduces dual branches: one processes spatial domain input through convolutional operations, while the other extracts and processes frequency-domain features via 2D Haar wavelet decomposition, convolutional operation, and inverse DWT reconstruction. These branches merge to create an integrated spatial-frequency representation, further refined through convolutional and diagonal Gaussian mapping into a robust latent representation. We utilize a new satellite imagery dataset housed by the TerraFly mapping system to validate our method. Experimental results across several performance metrics highlight the efficacy of the proposed method at enhancing latent space representation. 
\end{abstract}


\section{Introduction}
\label{sec:intro}
VAEs \cite{kingma2013auto} are generative autoencoder frameworks that have demonstrated effectiveness across diverse tasks, including image recognition \cite{gogna2019discriminative} and denoising \cite{majumdar2018blind}. A significant contribution of VAEs is their integration within diffusion models \cite{sohl2015deep}, leading to the development of LDMs \cite{rombach2022high}. Compared to traditional diffusion models operating purely in pixel space \cite{ho2020denoising, song2020denoising}, LDMs significantly reduce computational complexity by performing diffusion processes within a compressed latent space constructed by VAE.

Recent advancements in LDMs included Stable Diffusion versions, such as Stable Diffusion 1.5, Stable Diffusion XL (SDXL), and Stable Diffusion 3.5, demonstrating substantial success, particularly in text-to-image generation tasks. Moreover, LDM has increasingly been adopted in RS applications, including tasks such as text-to-image generation \cite{sebaq2024rsdiff}, image super-resolution \cite{luo2024satdiffmoe}, vegetation mapping \cite{zhao2024vegediff}, and large-scale image analysis \cite{khanna2023diffusionsat}.

A key factor contributing to the success of LDMs is the use of VAEs pretrained on general-purpose image datasets. However, satellite imagery differs significantly from general natural imagery because of unique spectral characteristics and limited temporal sampling. Therefore, relying on VAEs pretrained on unrelated datasets, such as those consisting of common natural objects, can result in suboptimal performance when applied to satellite imagery tasks. Despite the pivotal role autoencoders play in achieving high-quality image generation with reduced computational requirements, innovations specifically enhancing the latent space representation of VAEs for RS applications remain limited. Most existing research primarily focuses on improving backbone networks in LDMs rather than optimizing the intrinsic latent encoding mechanism.

To bridge this gap, the present study introduces a new approach by incorporating frequency-domain feature analysis through the 2D DWT into the VAE's encoding module. By leveraging the complementary strengths of spatial and frequency-domain representations, this study aims to enhance latent encoding performance in VAEs for satellite imagery data.
\begin{figure*}
  \centering 
  \includegraphics[width=0.96\linewidth]{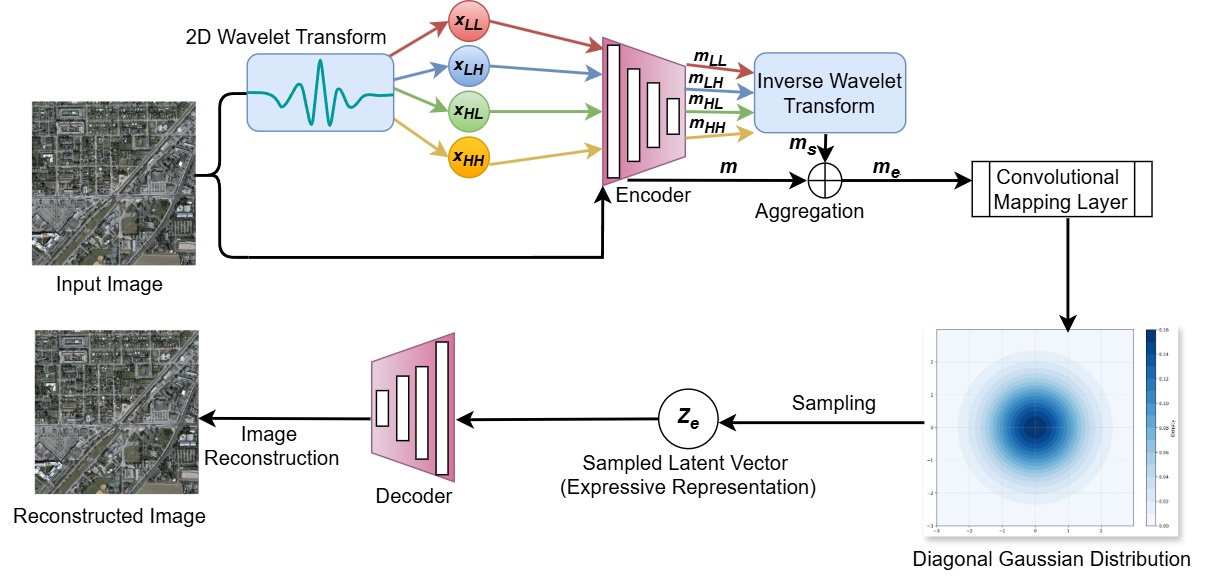}
  \caption{Overview of the proposed ExpDWT-VAE architecture. It extends the standard VAE by integrating a dual-branch encoding: one with 2D wavelet decomposition and another without. \(z_e\) represents the enhanced latent representation.}
  
  \label{fig:DWTVAE}
\end{figure*}

\section{Related Work}
While several studies have incorporated frequency-domain techniques such as Fast Fourier Transform (FFT) \cite{cooley1965algorithm} and DWT \cite{mallat1989theory} into diffusion models, explicit application within latent diffusion frameworks remains limited. For example, \cite{xiong2024rshazediff} applied FFT to denoising diffusion probabilistic models (DDPM) for image dehazing in Remote Sensing. \cite{phung2023wavelet} leveraged wavelet-based frequency methods to accelerate denoising steps in diffusion models, achieving success in general public datasets.\\
Similarly, Jiang et al. \cite{jiang2023low} integrated wavelet frequencies into both forward and backward diffusion processes for enhanced low-light imaging. However, these frequency-based techniques primarily targeted improvement in diffusion operations themselves rather than enhancing the intrinsic latent representations of autoencoders within latent diffusion models. Recently, LiteVAE \cite{sadat2024litevae} demonstrated the use of multi-level DWT for latent encoding within a VAE framework, effectively reducing model parameters. LiteVAE processes wavelet coefficients through separate feature extractors, subsequently combined via a U-Net aggregation module, achieving computational efficiency. However, LiteVAE does not explicitly focus on enhancing latent representation quality nor address satellite imagery applications.\\
In contrast, our work introduces a distinct approach by integrating 2D DWT from a single-level decomposition into the encoder of VAEs, using it as a facilitator to enhance latent encoding without altering the underlying architecture.

\section{Proposed Approach}
In this section, we first provide a brief overview of VAEs and their role in LDMs. Subsequently, we explore frequency-based properties through the DWT, and finally introduce our proposed enhancement in the latent encoding process.
\subsection{VAE in LDM and DWT} VAEs consist of two modules: an encoder \(\mathcal{E}\), which maps an input image $x \in \mathbb{R}^{H \times W \times 3}$ to a compressed latent representation $z = \mathcal{E}(x) \in \mathbb{R}^{h \times w \times c}$, and a decoder $D$ that reconstructs the image from the latent code, yielding $\tilde{x} = D(\mathcal{E}(x)) \approx x$. The downsampling factor $f$ determines the relationship between the spatial dimensions, given by $f = H/h = W/w$.
In the context of LDMs, VAEs function as perceptual compressors, ensuring that reconstructions remain faithful to the original images while significantly reducing the computational complexity associated with pixel-space diffusion. VAE training typically leverages reconstruction loss \( L_{\text{recon}} \), incorporating perceptual and pixel-level fidelity metrics \cite{zhang2018unreasonable}, along with regularization of latent representations via Kullback–Leibler divergence \cite{kingma2013auto}. Similarly, adversarial training \cite{goodfellow2014generative} is often incorporated by introducing a discriminator to distinguish between real and reconstructed images, facilitating the reconstruction of high-quality images. 
The 2D DWT \cite{mallat1989theory} provides a powerful multi-resolution analysis framework, enabling simultaneous spatial and frequency localization. DWT decomposes images into low-frequency approximation coefficients and high-frequency detail coefficients.

In our setup, the DWT is specifically implemented using the Haar wavelet at decomposition level 1, with symmetric padding applied to favor smooth decomposition and accurate reconstruction of features without loss of relevant information.

\subsection{ExpDWT-VAE}
The proposed method, ExpDWT-VAE, as depicted in Figure \ref{fig:DWTVAE}, integrates DWT into the VAE architecture to enhance latent representations specifically for satellite imagery. Given an input image \( x \), the encoder \( \mathcal{E} \) initially processes it through multiple convolutional and downsampling layers to generate a spatial-domain feature map \( m \). Concurrently, we apply the above-defined 2D DWT, decomposing the image into one low-pass approximation and three high-pass detail sub-bands: \( x \rightarrow \{x_{LL}, x_{LH}, x_{HL}, x_{HH}\} \). As illustrated by the differently colored arrows in Figure \ref{fig:DWTVAE}, these four frequency-domain sub-bands individually undergo the same convolutional encoding operations used by the main encoder, producing frequency-domain feature maps, \(m_{LL}\), \(m_{LH}\), \(m_{HL}\), \(m_{HH}\).

Subsequently, the inverse DWT is carefully applied to these encoded frequency-domain feature maps to reconstruct a unified spatial-domain feature map \( m_s \). The spatial feature \( m \) from the original encoder pathway is then element-wise added to \( m_s \), yielding an enriched spatial-frequency aware feature map:
\begin{equation}
m_e = m + m_s.
\end{equation}
The combined feature map \( m_e \) is passed through a convolutional mapping layer \( \mathcal{Q} \), which computes the parameters of a diagonal Gaussian distribution, the mean \( \mu \) and the log-variance \( \log\sigma^2 \):
\begin{equation}
(\mu, \log\sigma^2) = \mathcal{Q}(m_e).
\end{equation}
This parameterization defines the posterior distribution \( q(z_e \mid m_e) \), from which the latent representation \( z_e \) is sampled using the reparameterization trick.
Here, \( z_e \) serves as the expressive latent representation, obtained through iterative integration of DWT and the inverse DWT by effectively combining spatial and frequency-domain information.

\section{Experiments}
\label{sec:formatting}
For computational evaluation, satellite (raster) images with RGB channels were collected from the TerraFly mapping system using a Python-based web scraping procedure. Data were downloaded from three U.S. states: Florida, Georgia, and Texas, with a zoom level of 15 to maintain consistent ground-level resolution. After preprocessing and refinement, a total of 6,000 images were selected for training and validation. We name this dataset TerraFly-Sat.

We evaluated the models using Variance ($\sigma^2$) to assess latent space expressiveness, LPIPS for perceptual similarity, PSNR and SSIM for reconstruction quality, and FID and KID for distribution alignment. Experiments were conducted on two latent dimensions: 32$\times$32$\times$4 and 64$\times$64$\times$3, using PyTorch 3.9.20 on NVIDIA A100-PCI GPUs with 80 GB HBM2 memory.

\subsection{Results}
\begin{figure}[!htbp]
  \centering
  \includegraphics[width=1\linewidth]{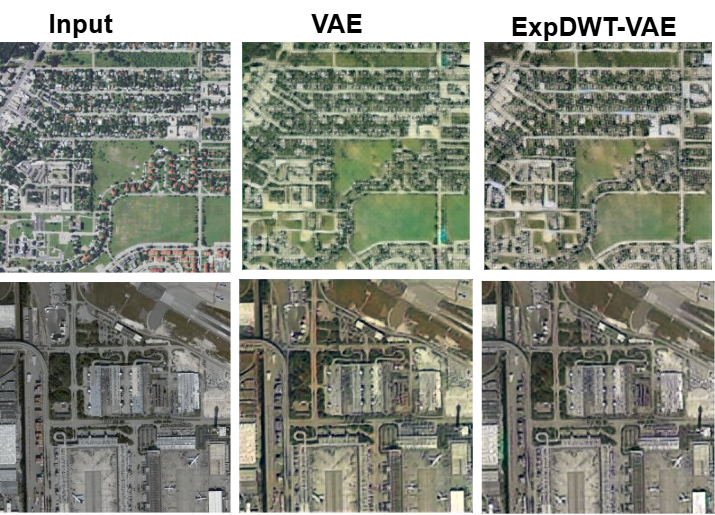}
  \caption{Visual comparison of input images and their reconstructions by VAE and ExpDWT-VAE on the TerraFly-Sat dataset.}
  \label{fig:Reconstruction_Image_Illustration}
\end{figure}

Table \ref{tab:vae-comparison} summarizes the quantitative results, with visual examples in Figure \ref{fig:Reconstruction_Image_Illustration}. Notably, ExpDWT-VAE achieves higher latent variance, supporting our goal of enriching latent space representation. Since the decoder remained unchanged, improvements in reconstruction metrics (LPIPS, PSNR, SSIM, FID and KID) suggest the entire VAE pipeline advancement from our encoder design. This highlights the method’s potential for real-world satellite image modeling.

\begin{figure}[!htbp]
  \centering
  \includegraphics[width=1\linewidth]{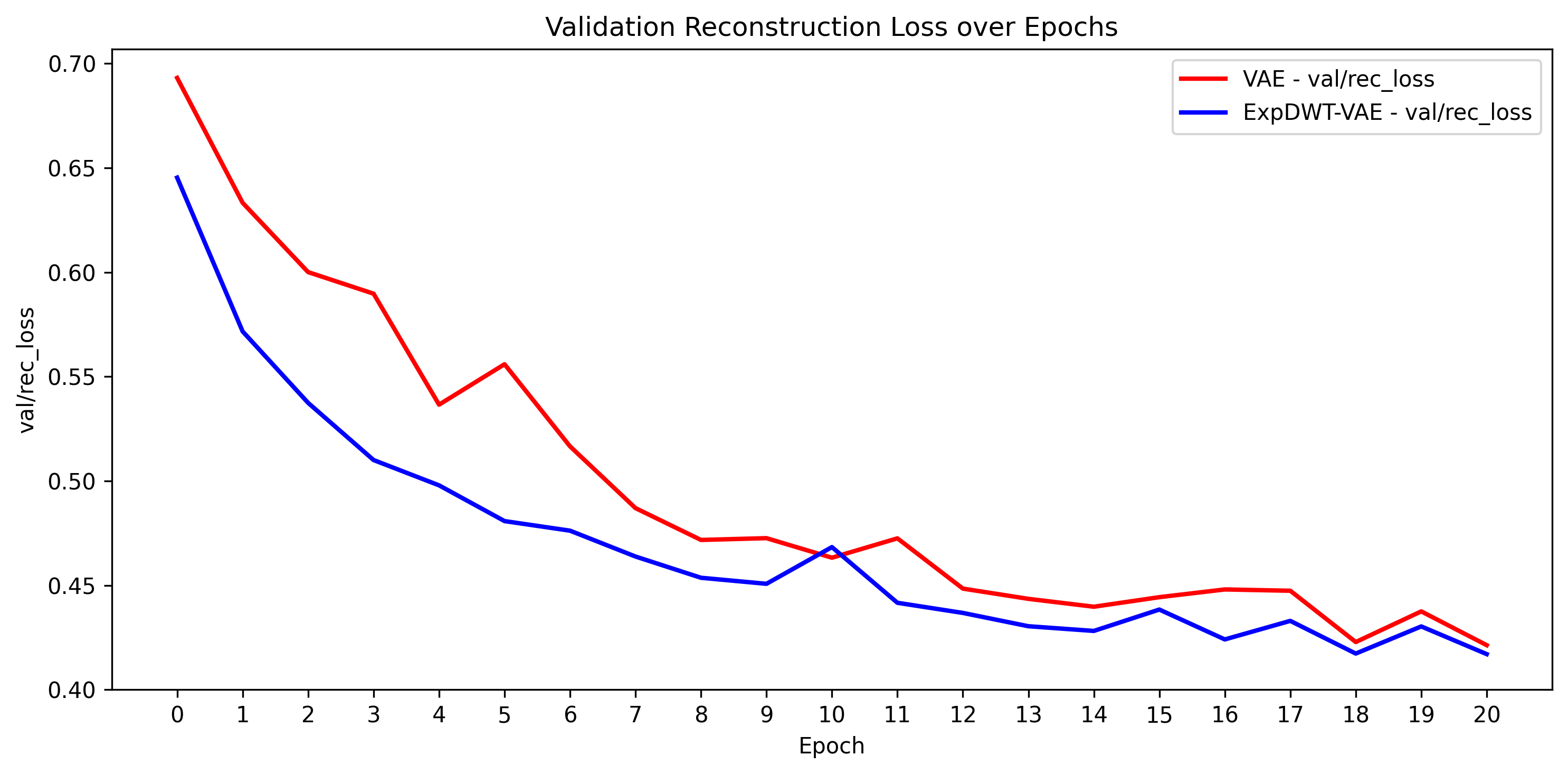}
  \caption{Validation Reconstruction Loss Obtained from ExpDWT-VAE and VAE.}
  \label{fig:Reconstruction_Loss_Plot}
\end{figure}

Figure \ref{fig:Reconstruction_Loss_Plot} shows that ExpDWT-VAE starts with lower validation loss (0.64 vs. 0.69) and maintains an edge throughout training. Importantly, these gains were achieved without any hyperparameter tuning. We hypothesize that tailoring hyperparameters for frequency-domain processing could further enhance performance, an avenue for future work. The code is available at \href{https://github.com/amaha7984/ExpDWT-VAE} {https://github.com/amaha7984/ExpDWT-VAE}.

\begin{table*}
  \centering
  \begin{tabular}{@{}llcccccc@{}}
    \toprule
    Latent Dim & Model & Variance ($\sigma^2$) & LPIPS & PSNR & SSIM & FID & KID \\
    \midrule
    \multirow{2}{*}{32$\times$32$\times$4}
      & VAE           & 3.41 & 0.31  & 19.22 & \textbf{0.4984} & 64.44 & 0.0299 ± 0.0005 \\
      & ExpDWT-VAE    & \textbf{6.92} & \textbf{0.27}  & \textbf{19.76} & 0.4963 & \textbf{62.07} & \textbf{0.0271 ± 0.0006} \\
    \multirow{2}{*}{64$\times$64$\times$3}
      & VAE           & 3.70     & 0.25     & 21.24     & 0.73    & 60.22    & 0.0287 ± 0.0005              \\
      & ExpDWT-VAE    & \textbf{8.95} & \textbf{0.21} & \textbf{22.80} & \textbf{0.74} & \textbf{41.30} & \textbf{0.0146 ± 0.0004} \\
    \bottomrule
  \end{tabular}
  \caption{Comparative performance of VAE and ExpDWT-VAE across different latent dimensions on the TerraFly-Sat dataset.}
  \label{tab:vae-comparison}
\end{table*}

\section{Conclusion}
While VAEs significantly reduce computational complexity in latent diffusion models, existing studies rarely focus on refining the latent space itself. In this work, we proposed ExpDWT-VAE, a model that integrates 2D Haar Discrete Wavelet Transform (DWT) into the encoder of a VAE. This integration enables frequency-aware refinement alongside spatial encoding, resulting in more expressive latent representations. We introduced a new satellite imagery dataset, TerraFly-Sat, and presented both quantitative and qualitative evaluations that demonstrate the effectiveness of our approach over the standard VAE.


\section*{Acknowledgments}
This material is based in part upon work supported by the National Science Foundation under Grant Nos. CNS-2018611 and FDEP grant C-2104.

\clearpage
\renewcommand{\refname}{References}
\bibliographystyle{unsrt}

\end{document}